\documentclass[11pt,a4paper]{article}
\usepackage[hyperref]{eacl2021}
\usepackage{times}
\usepackage{latexsym}

\usepackage{microtype}

\usepackage{graphicx}
\usepackage{paralist}
\usepackage{booktabs}
\usepackage{multirow}
\usepackage{amsmath, amsthm, amsfonts, amssymb, amscd, bm}
\usepackage{subcaption}
\usepackage{comment}
\usepackage{enumitem}
\usepackage{wrapfig}
\usepackage{tabularx}

\usepackage{listings}
\usepackage{color}

\ifx \@etal \@empty  \fi
\ifx \@eg \@empty \newcommand{\eg}{e.g.,~} \fi
\ifx \@ie \@empty \newcommand{\ie}{i.e.,~} \fi
\ifx \@etc \@empty \newcommand{\etc}{etc.} \fi

\newcommand{\bb}{\boldsymbol{b}}
\newcommand{\bc}{\boldsymbol{c}}

\newcommand{\be}{\boldsymbol{e}}

\newcommand{\bv}{{\boldsymbol{v}}}
\newcommand{\bw}{{\boldsymbol{w}}}

\newcommand{\bV}{{\boldsymbol{V}}}
\newcommand{\bW}{{\boldsymbol{W}}}

\newcommand{\onedot}{.\,}
\newcommand{\eg}{\emph{e.g}\onedot}
\newcommand{\ie}{\emph{i.e}\onedot}
\newcommand{\etc}{\emph{etc}\onedot}

\newcommand{\fig}{Fig.~}
\newcommand{\eq}{Eq.\,}

\newcommand{\sect}{Section~}
\newcommand{\tab}{Table~}
\newcommand{\ap}{Appendix~}
\newcommand\closedots{\makebox[1em][c]{.\hfil.\hfil.}}
\newcommand{\ra}[1]{\renewcommand{\arraystretch}{#1}} 

\graphicspath{ {./figures/} }

\aclfinalcopy 

\title{Reasoning over Vision and Language:\\Exploring the Benefits of Supplemental Knowledge}

\author{Violetta Shevchenko, Damien Teney, Anthony Dick, Anton van den Hengel \\
The University of Adelaide\\
{\tt\small \{violetta.shevchenko,damien.teney,anthony.dick,anton.vandenhengel\}@adelaide.edu.au}
}

\begin{document}
\maketitle
\begin{abstract} 
The limits of applicability of vision-and-language models are defined by the coverage of their training data.
Tasks like vision question answering (VQA) often require commonsense and factual information beyond what can be learned from task-specific datasets.
This paper investigates the injection of knowledge from general-purpose knowledge bases (KBs) into vision-and-language transformers.
We use an auxiliary training objective that encourages the learned representations to align with graph embeddings of matching entities in a KB.
We empirically study the relevance of various KBs to multiple tasks and benchmarks.
The technique brings clear benefits to knowledge-demanding question answering tasks (OK-VQA, FVQA) by capturing semantic and relational knowledge absent from existing models.
More surprisingly, the technique also benefits visual reasoning tasks (NLVR2, SNLI-VE).
We perform probing experiments and show that the injection of additional knowledge regularizes the space of embeddings, which improves the representation of lexical and semantic similarities.
The technique is model-agnostic and can expand the applicability of any vision-and-language transformer with minimal computational overhead.
\end{abstract}



\section{Introduction}
\begin{figure}[t]
	\centering
	\includegraphics[width=0.9\columnwidth]{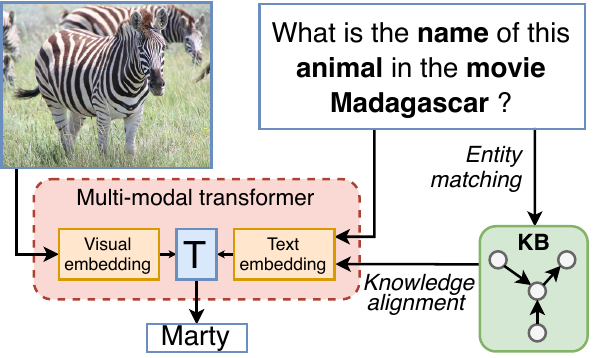}
	\caption{We inject additional information from knowledge bases (KBs) in a vision-and-language transformer.
	We first preprocess the KB into knowledge embeddings.
	When training the transformer, we use an auxiliary objective that aligns its learned representations with knowledge embeddings.
	In the context of visual question answering (pictured), the model is able to answer questions that require knowledge that is typically not captured in task-specific datasets.}
	\label{fig:scheme}
\end{figure}

The last few years have seen a surge of interest for vision and language (V\&L) tasks. These include visual question answering for example (VQA), in which the machine must answer a question about an image (\fig\ref{fig:scheme}). V\&L tasks require processing two modalities and reasoning over textual, visual, and abstract concepts.

The current state of the art in V\&L are models based on transformers such as BERT~\cite{devlin2018bert} that have been extended to handle visual inputs~(\eg~\citet{tan2019lxmert}).
These models are usually pretrained on a collection of datasets of paired textual and visual data.
Suitable datasets include image captioning data~\cite{chen2015microsoft,sharma2018conceptual} and VQA data~\cite{goyal2017making,hudson2019gqa}.
Despite their large scale, these datasets only cover limited domains.
Many are based on images from COCO~\cite{lin2014microsoft} and Visual Genome~\cite{krishna2017visual} and the linguistic diversity of textual annotation is limited.
The V\&L tasks that we are ultimately interested in require knowledge beyond current datasets, \eg about specific events, named entities, common sense, and abstract concepts~(see \fig\ref{fig:examples}).

This paper focuses on the expansion of the applicability of V\&L models with additional knowledge.
During training, we infuse the model with knowledge from an external source, distinct from datasets of paired V\&L data.
The challenge is that standard V\&L data is not annotated or paired with such additional knowledge.
Even though techniques have been proposed to exploit additional data in NLP, including text-based question answering~\cite{kafle2019overview,lv2019graph,lin2019kagnet,rajani2019explain},
little work has been done on the extension to V\&L.
Works in NLP with benchmarks of knowledge-demanding questions~\cite{yang2015wikiqa,talmor2018commonsenseqa,clark2018think,mihaylov2018can,sap2019socialiqa} have shown that knowledge bases (KBs) contain information that can benefit large-scale pretrained models.
Motivated by this line of evidence, we aim to evaluate similar mechanisms for V\&L tasks.

In the context of VQA, datasets of knowledge-demanding questions~\cite{wang2017fvqa,marino2019okvqa,shah2019kvqa} have proved challenging for existing methods.
For example, a question like \textit{Is the man eating healthy food~?} requires recognizing a type of food and relating it to its nutritional quality.
Learning this type of knowledge from VQA training examples would be clearly inefficient.
Other examples of challenging questions involve references to named entities such as brands, locations, or movie titles.
For example, the question \textit{Levi's is a popular brand of what item shown here~?} requires knowledge of the brand's specialization before identifying the correct element in the image.
We believe that embedding this type of knowledge in a model is a necessary step to enable progress on complex multimodal question answering.

This paper describes a technique to inject information from KBs into a transformer-based model during its training~(see \fig\ref{fig:scheme}).
We take inspiration from~\citet{goodwin2019bridging} and adapt their regularizer from text-based models to V\&L transformers.
We provide an implementation of our method on top of the popular LXMERT model~\cite{tan2019lxmert}. We use it to investigate the suitability of several KBs to different tasks and benchmarks.

\noindent
Our contributions are summarized as follows.
\setlist{leftmargin=*}
\begin{itemize}
	\item We describe a method to inject information from knowledge bases during the training of vision-and-language transformers.
	\item We implement the method on top of the popular LXMERT model with the ConceptNet and Wikidata knowledge bases.
	\item We perform an extensive empirical evaluation on four downstream tasks. We demonstrate clear improvements on knowledge-demanding VQA (OK-VQA and FVQA datasets) and visual reasoning (NLVR2 and SNLI-VE datasets).
	\item We conduct an in-depth analysis, including ablations and probing experiments.
	They show that we improve the representation of lexical, semantic, and relational knowledge that is lacking in typical V\&L models.
	This explains the surprising improvements on tasks that do not explicitly depend on external knowledge.
\end{itemize}

\section{Related Work}
\label{sec:related-work}

\paragraph{Vision-and-language tasks} require joint processing of visual and textual data \eg for image captioning~\cite{anderson2018bottomup,hossain2019comprehensive} and VQA~\cite{wu2016survey,teney2017spm}.
They have historically been approached with task-specific models, but recent transformer-based models~\cite{vaswani2017attention} were shown to be applicable to a variety of tasks.
A transformer can thus be pretrained on multiple datasets and then fine-tuned for one specific task~\cite{tan2019lxmert,zhou2019unified,alberti2019fusion,chen2019uniter,li2019unicodervl,lu2019vilbert,su2019vlbert,li2019visualbert,sun2019videobert}.
This paper describes a method to embed additional information in a transformer-based model (``additional'' to the pretraining and fine-tuning datasets). Our implementation builds on the popular LXMERT model~\cite{tan2019lxmert}.

\paragraph{Additional knowledge in NLP} can help with tasks requiring commonsense or factual information~\cite{storks2019commonsense}.
Various techniques have been proposed to improve transformers such as BERT~\cite{devlin2018bert}.
\citet{zhang2019ernie} proposed ERNIE, which feeds graph embeddings of text entities to the model.
\citet{peters2019knowledgea} proposed KnowBert, a similar technique suitable to multiple KBs.
\citet{levine2019sensebert} used WordNet to aid in the masked-word prediction objective, and improve lexical understanding in downstream tasks.
\citet{ye2019align} proposed a multiple-choice question answering pretraining task and improved performance on multiple datasets requiring commonsense reasoning.
\citet{liu2019k} addressed noise issues by controlling the amount of domain-specific knowledge infused into the model.
\citet{goodwin2019bridging} proposed OSCAR, a regularization method to inject ontological knowledge in a pretrained language model.
\citet{wang2019kepler} proposed to simultaneously learn knowledge representations while optimizing a masked-language objective, rather than using pretrained knowledge embeddings.
All of these works were applied to NLP tasks. This paper studies the suitability of similar mechanisms to V\&L tasks by applying the technique of \citet{goodwin2019bridging} to a multimodal transformer.

\paragraph{Knowledge-based VQA} refers to benchmarks designed to require additional information~\cite{marino2019okvqa,shah2019kvqa,wang2015explicit,wang2017fvqa}.
Models have been proposed that retrieve such information from KBs based on question and image contents~\cite{wang2015explicit,wu2016ask,wang2017fvqa,narasimhan2018straight}.
Some works use other sources of external information at training~\cite{teney2016zsvqa} or test time~\cite{teney2017visual,teney2019actively} but they showed limited improvements on VQA benchmarks.
Recently proposed ConceptBERT~\cite{garderes2020conceptbert} model jointly learns visual, textual and knowledge embeddings to fuse commonsense information into VQA models. 
This paper describes a method applicable to a variety of sources of information and to tasks beyond VQA. We also show that different tasks benefit from different types of information.

\section{Methodology}

\begin{figure*}[t]
	\centering
	\includegraphics[width=0.9\linewidth]{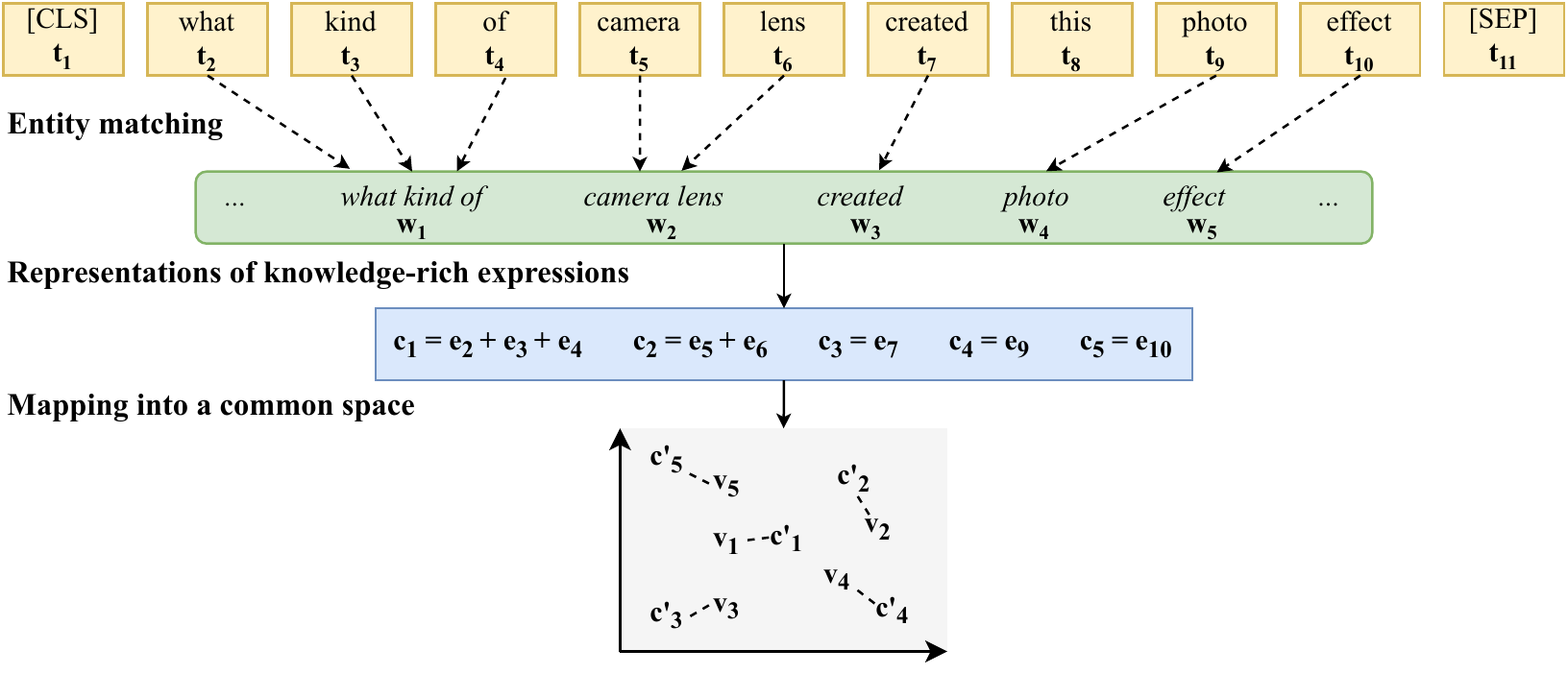}
	\caption{Summary of the approach.
	During training, we first match tokens in the V\&L training data (yellow) with entities of the knowledge base (green).
	We then train the transformer with an additional loss to align the learned representations $\bc_k$ (sums of word embeddings in knowledge-rich expressions from the transformer, in blue) with the knowledge embeddings $\bv_k$ derived from the knowledge base.}
	\label{fig:approach}
\end{figure*}

We describe a general, simple yet effective technique to embed additional knowledge into a transformer-based model.
It is compatible with existing multimodal transformers and thus suitable to a variety of tasks.
The method proceeds in three stages (see \fig\ref{fig:approach}).
\setlist{nolistsep,leftmargin=*,itemsep=2pt,topsep=0pt}
\begin{enumerate}
	\item We preprocess the additional knowledge into a set of vector representations that we call knowledge embeddings.
	For example, with a relational knowledge base, we apply a graph embedding method to obtain a vector representation of every of its entities.
	Each is associated with a textual expression. 
	\item We match sentences in the V\&L training data with knowledge embeddings. Matches in the training data are referred to as knowledge-rich expressions.
	\item During the training of the transformer (pretraining and/or fine-tuning), we optimize an additional objective that aligns its learned representations of knowledge-rich expressions with the matching knowledge embeddings.
\end{enumerate}
We now describe each stage in details.

\subsection{Representations of Additional Knowledge}
The versatility of the approach rests on storing the additional knowledge as a set of knowledge embeddings $\bV=\{\bv_i\}_i$ with $\bv_i \in \mathbb{R}^{d_v}$.
These can be produced by preprocessing sources such as text corpora (with word embedding methods) 
or relational knowledge bases (with a graph embedding method, see \sect\ref{secKbs}).
These knowledge embeddings capture semantic information about the relations between entities, which we will incorporate into the V\&L model.
Each knowledge embedding $\bv_i$ is associated with an instantiation $\bw_i$ in the V\&L data.
In this work, the $\bw_i$ are purely textual (single- and multiple-word expressions) but future work could consider visual representations of concepts represented by knowledge embeddings.
We denote with $\bW=\{\bw_i\}_i$ the vocabulary of these instantiations.

\subsection{Matching V\&L Training Data with Knowledge Embeddings}
We match parts of the training data with entities having additional knowledge.
Since our vocabulary $\bW$ contains textual expressions, we use greedy longest-string matching to identify subsequences in the data that match any $\bw_i$.
Possible improvements left for future work include performing named entity recognition and homonym disambiguation.
In practice, entities in our KBs have unique textual representations, making homonyms a non-issue.

We represent text in the training data as a sequence of tokens $T=(t_1, ...\,, t_M)$.
The transformer internally maps each token $t_j$ to a word embedding $\be_j \in \mathbb{R}^{d_e}$.
Each correspondence identified by the matching algorithm is of the form of a subsequence $(t_{a_k}, ...\,, t_{b_k}) \subset T$ and $\bw_k \in \bW$.
We refer to such a subsequence as a \emph{knowledge-rich expression}.
To obtain a fixed-size representation $\bc_k$ of a knowledge-rich expression, we sum the word embeddings of its constituent tokens \ie~ $\bc_k = \be_{a_k} + \closedots + \be_{b_k}$.

\begin{table*}[t!]
	\ra{1.15}
	\begin{center}
	\renewcommand\tabcolsep{15pt}
	\footnotesize
	\begin{tabularx}{\linewidth}{Xcccc}
		 \toprule
		 & OK-VQA & FVQA & SNLI-VE & NLVR2\\\midrule
		 \multicolumn{5}{l}{\hspace{-8pt}{Pretraining with COCO captions, Visual Genome captions, VQA v2, GQA}\vspace{3pt}}\\

		 Baseline (LXMERT) & 37.26 $\pm$ 0.23 & 52.30 $\pm$ 0.09 & 74.05 $\pm$ 0.19 & 71.31 $\pm$ 0.56\vspace{4pt}\\

		 With ConceptNet PT & \textbf{39.04 $\pm$ 0.24} & 54.08 $\pm$ 0.09 & 75.18 $\pm$ 0.21 & 71.61 $\pm$ 0.24\\
		 With ConceptNet FT & 36.99 $\pm$ 0.01 & 52.19 $\pm$ 0.15 & 74.80 $\pm$ 0.08 & 70.82 $\pm$ 0.34\\
		 With ConceptNet PT+FT & 38.56 $\pm$ 0.31 & \textbf{54.27 $\pm$ 0.28} & \textbf{75.24 $\pm$ 0.22} & \textbf{72.59 $\pm$ 0.23}\vspace{4pt}\\\midrule

		 \multicolumn{5}{l}{\hspace{-8pt}{Pretraining with VQA v2, GQA}}\vspace{3pt}\\

		 Baseline (LXMERT) & 36.71 $\pm$ 0.22 & 50.38 $\pm$ 0.24 & 73.57 $\pm$ 0.17 & 67.88 $\pm$ 0.66\vspace{4pt}\\

		 With ConceptNet PT & 38.05 $\pm$ 0.41 & \underline{51.94 $\pm$ 0.25} & 74.07 $\pm$ 0.48 & 69.47 $\pm$ 0.20\\
		 With ConceptNet FT & 36.73 $\pm$ 0.50 & 50.54 $\pm$ 0.08 & 74.24 $\pm$ 0.15 & 67.09 $\pm$ 0.32\\
		 With ConceptNet PT+FT & \underline{38.12 $\pm$ 0.11} & 51.53 $\pm$ 0.17 & \underline{74.26 $\pm$ 0.22} & \underline{69.69 $\pm$ 0.12}\vspace{4pt}\\

		 With Wikidata PT & 37.39 $\pm$ 0.35 & 51.00 $\pm$ 0.03 & 73.48 $\pm$ 0.17 & 68.27 $\pm$ 0.89\\
		 With Wikidata FT & 36.31	$\pm$ 0.28 & 50.59 $\pm$ 0.24 & 73.60 $\pm$ 0.26 & 67.64 $\pm$ 0.19\\
		 With Wikidata PT+FT & 37.43 $\pm$ 0.25 & 50.74 $\pm$ 0.29 & 73.50 $\pm$ 0.04 & 68.25 $\pm$ 0.51\\
		 \bottomrule
	\end{tabularx}
	\caption{Overall results. Our model with ConceptNet during {p}re{t}raining and {f}ine-{t}uning (ConceptNet PT+FT) generally proves best. We report the average accuracy (\%) $\pm$ one standard deviation over three random seeds.}
	\label{tab:overall-acc}
	\end{center}
\end{table*}

\subsection{Aligning Learned Representations with Knowledge Embeddings}
The core of the method is an additional training objective that encourages the transformer to produce representations of knowledge-rich expressions ($\bc_k$) that collectively align with knowledge vectors ($\bv_k$).
A vector $\bc_k$ as defined above is the representation learned by the model of a knowledge-rich expression.
We do not expect these vectors to correspond to their matching knowledge embeddings $\bv$, but we desire them to capture the information globally represented in the \emph{relations} between the knowledge embeddings in $\bV$.
Therefore, we define a new linear layer that maps a learned representation $\bc_k \in \mathbb{R}^{d_e}$ to $\bc'_k \in \mathbb{R}^{d_v}$:
\begin{equation}
	\label{eq:embKnowledgeExpression}
	\bc'_k = \bW_{c} \bc_k \,+\, \bb_{c}
\end{equation}
where $\bW_{c} \in \mathbb{R}^{d_e \times d_v}$ and $\bb_{c} \in \mathbb{R}^{d_v}$ are learned weights and biases.
We then define our alignment loss that encourages each projection $\bc'_k$ to be close to its corresponding knowledge embedding $\bv_k$:
\begin{equation}
	\label{eq:loss}
	L_\textrm{align} = \Sigma_k \| \bc'_k \,-\, \bv_k\|^2 ~~.
\end{equation}

Together, \eq\ref{eq:embKnowledgeExpression} and \ref{eq:loss} encourage the global structure of the learned representations $\bc$ to align with the set of knowledge embeddings $\bV$ through the projection $\bW_{c}$.
The learned representations incorporate information from the knowledge embeddings while allowing the transformer to also represent task-specific information.
The model is trained for the combination of its original main loss, and the new alignment loss weighted by a hyperparameter $\lambda$ \ie $L = L_\textrm{main} + \lambda \, L_\textrm{align}$.

\subsection{Training Strategies}
We experimented with the application of the method during {p}re{t}raining and/or {f}ine-{t}uning, referred to as FT, PT, and PT+FT below.
Enabling the additional objective during pretraining can benefit from the larger overlap between the training data and the additional knowledge.
However, most pretraining tasks do not specifically require additional knowledge, and the model may not learn to effectively use it.
During fine-tuning on knowledge-demanding tasks, the model is likely to better learn to capture and use relevant additional knowledge.

\subsection{Implementation}
\label{sec:implementation}
We implemented the method on top of LXMERT~\cite{tan2019lxmert}, the state-of-the-art model on multiple tasks at the onset of this project. This model is pretrained on five captioning and VQA datasets: COCO and Visual Genome captions, VQA v2~\cite{goyal2017making}, GQA~\cite{hudson2019gqa} and Visual Genome QA~\cite{zhu2016visual7w}.
We also include experiments with scaled-down pretraining on two datasets (VQA v2 and GQA) which will ease the computational cost of replication (details in \ap\ref{appendix:implem}).

\section{Experiments}
\begin{table*}[t]
	\ra{1.15}
	\begin{center}
	\renewcommand\tabcolsep{5.5pt}
	\footnotesize
	\begin{tabularx}{\linewidth}{Xccccccccccc}
		 \toprule
		 & \multicolumn{11}{c}{OK-VQA}\\
		 \cmidrule{2-12}
		 & VT & BCP & OMC & SR & CF & GHLC & PEL & PA & ST & WC & Other\\
		 \midrule
		 Baseline (LXMERT) & 34.91 & 29.34 & 35.22 & 46.41 & 40.16 & 32.58 & 32.68 & 35.75 & \textbf{35.79} & \underline{48.68} & 35.96\\
		 \midrule
		 With ConceptNet PT & \textbf{37.40} & \textbf{31.94} & \underline{37.66} & \textbf{47.52} & \underline{40.82} & \textbf{38.39} & \textbf{35.06} & \textbf{37.20} & 31.98 & 48.63 & \textbf{38.42}\\
		 With ConceptNet FT & 34.46 & 29.03 & 35.89 & 46.02 & 39.73 & 33.66 & 32.57 & 35.51 & \underline{35.16} & 45.84 & 35.75\\
		 With ConceptNet PT+FT & \underline{37.11} & \underline{31.47} & \textbf{39.06} & \underline{46.52} & \textbf{40.89} & \underline{36.50} & \underline{34.08} & \underline{36.16} & 33.73 & \textbf{49.56} & \underline{37.05}\\
		 \bottomrule
	\end{tabularx}
	\caption{Accuracy (\%) on OK-VQA per question category (description in the Appendix, \tab\ref{tab:qtypes}). The method obtains clear improvements on categories related to commonsense knowledge such as \textbf{BCP}: brands, companies and products, \textbf{OMC}: objects, material and clothing, \textbf{PEL}: people and everyday life, \textbf{VT}: vehicles and transportation.}
	\label{tab:okvqa-acc}
	\end{center}
\end{table*}

We performed a suite of experiments with multiple tasks and benchmarks.
Our objective is to evaluate the suitability of two popular knowledge bases (ConceptNet and Wikidata) to these tasks.

\paragraph{Datasets.}
We used four datasets.
The OK-VQA~\cite{marino2019okvqa} and FVQA~\cite{wang2017fvqa}
are two datasets for VQA with questions that require a wide range of commonsense and factual knowledge.
The SNLI-VE~\cite{xie2019visual} is a dataset for visual entailment.
Each instance involves an image ``premise'' and text ``hypothesis''. The task is to determine whether the image entails or contradicts the text, or whether they are neutral to each other.
The NLVR2 dataset~\cite{suhr2018corpusa} evaluates visual reasoning.
Each instance is a pair of images and a text description. The task is to determine whether the text accurately describes the pair of images.
See \ap\ref{appendix:datasets} for a comparison of the datasets.

\paragraph{Knowledge bases.}
\label{secKbs}
We experimented with two popular KBs.
\textbf{ConceptNet}~\cite{speer2017conceptnet} is a knowledge graph that encodes the meaning of expressions useful for general language understanding.
This decades-old project is built on a number of crowd-sourced and curated sources including dictionaries, encyclopedias, and ontologies.
We use the 300-dimensional Numberbatch embeddings distributed by the authors of ConceptNet.
They are built using the technique of retrofitting~\cite{faruqui2014retrofitting} to combine relational information from the KB with distributional semantics from word2vec~\cite{mikolov2013efficient}, GloVe~\cite{pennington2014glove}, and OpenSubtitles~\cite{tiedemann2012parallel}.
We only use the subset of English expressions ($\sim$500,000 entities).
\textbf{Wikidata}~\cite{wikidata} is a collaboratively edited database of general knowledge.
While ConceptNet mostly covers commonsense knowledge, Wikidata spans a larger domain, including historical events, celebrities, locations, science facts, \etc
We obtain 200-dimensional embeddings with the PyTorch-BigGraph graph embedding method~\cite{lerer2019pytorchbiggraph}. 
We use a subset of entities that have links to meaningful Wikipedia pages as done in~\cite{wang2019kepler}.
We also discard entities associated with stop words (\eg~\textit{the}, \textit{are}, \textit{there}) which are common in VQA questions but carry no important semantic information.
We retain $\sim$4.7M entities associated with $\sim$10M aliases.



\subsection{Overall Results}
We report the overall accuracy on all datasets in \tab\ref{tab:overall-acc}. Our approach clearly outperforms the baseline, with a \textbf{higher accuracy} on the knowledge-demanding VQA datasets OK-VQA (+1.78\%) and FVQA (+1.97\%). We also get clear improvements on the visual reasoning datasets SNLI-VE (+1.19\%) and NLVR2 (+1.28\%). These do not explicitly require  specific knowledge, so we hypothesized that the improvement is due to the richer linguistic representations learned by our model. We verified this hypothesis through probing experiments with the SentEval toolkit~\cite{conneau2018senteval}, which showed that our model better captures multiple semantic and syntactic properties of words (details in \ap\ref{appendix:probe}).

We provide example results in \fig\ref{fig:examples} and in \ap\ref{appendix:results}. 

The best training strategy is generally to use the additional objective \textbf{during both pretraining and fine-tuning} (PT+FT). Only on OK-VQA did the PT strategy perform slightly better.
Comparing PT alone with FT alone (the former being superior on all datasets) shows that fine-tuning the representations is not sufficient. The additional objective induces a globally different organization of the embedding space, which cannot be reorganized during fine-tuning alone. 
It is also interesting to note that the knowledge injection during pretraining is effective despite the pretraining tasks not specifically requiring additional knowledge.




\subsection{Nearest Neighbors in Embedding Space}
To better understand how the regularization with additional knowledge changes the structure of the learned embedding space, we examine nearest neighbors of embeddings of individual words (see \tab\ref{tab:embeds} and \ap\ref{appendix:embspace}).
Our model clearly captures more lexical and semantic information.
For example, the nearest neighbors for \textit{vitamin} are \textit{calcium} and \textit{supplements} with our model, but they are \textit{margarita} and \textit{amphibious} with the baseline. This stark qualitative difference was observed across the board and is not confined to cherry-picked examples.
The lack of linguistic information encoded by baseline V\&L transformers is unsurprising since their training data has limited linguistic diversity. 
Initializing a V\&L model with a pretrained BERT has been proposed to address this deficiency, but it was reported to give lower downstream performance by the authors of LXMERT.
In comparison, our method brings linguistic information while improving downstream performance.

\begin{table*}[h!]
	\ra{1.1}
	\begin{center}
	\renewcommand\tabcolsep{13pt}
	\footnotesize
	\begin{tabular}{@{}lll@{}}
		\toprule
		Word & Nearest neighbors with the baseline & Nearest neighbors with our model\\
		\midrule
		argentina & questioned, [PAD], neutron, dorset & argentine, uruguay, paraguay, mendoza \\
		behaviour & [PAD], absurd, authoritative, mba & behavior, behaviors, demeanor, behavioral \\
		bowling & boxing, smashed, dancing, 75 & bowler, bowled, cricket, tennis \\
		cottage & farmhouse, scan, condo, \#\#tta & cottages, bungalow, farmhouse, \#\#ode \\
		facebook & jade, brady, institution, utrecht & myspace, twitter, youtube, dit \\
		genes & [PAD], oro, subsistence, \#\#vah & gene, genetic, genetics, genome \\
		lecturer & greenberg, [unused983], avoidance, \#\#mour & professor, prof, professors, lectures \\
		playstation & splendid, indo, financial, tapping & xbox, wii, sega, consoles \\
		\bottomrule
	\end{tabular}
	\caption{Examples of nearest neighbors in the space of word embeddings learned by the baseline and by our model. Our model better captures lexical and semantic similarity.}
	\label{tab:embeds}
	\end{center}
\end{table*}

\begin{figure*}[t!]
	\centering
	\begin{subfigure}[t]{0.49\textwidth}
		\centering\includegraphics[width=\textwidth]{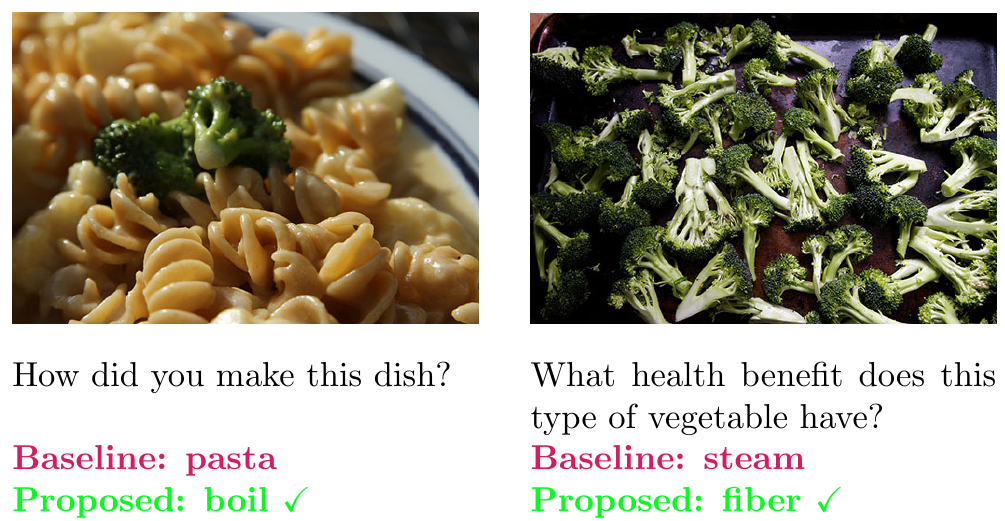}
		\caption{OK-VQA.\label{fig:ok-vqa}}
	\end{subfigure}
	\hfill
	\begin{subfigure}[t]{0.49\textwidth}
		\centering\includegraphics[width=\textwidth]{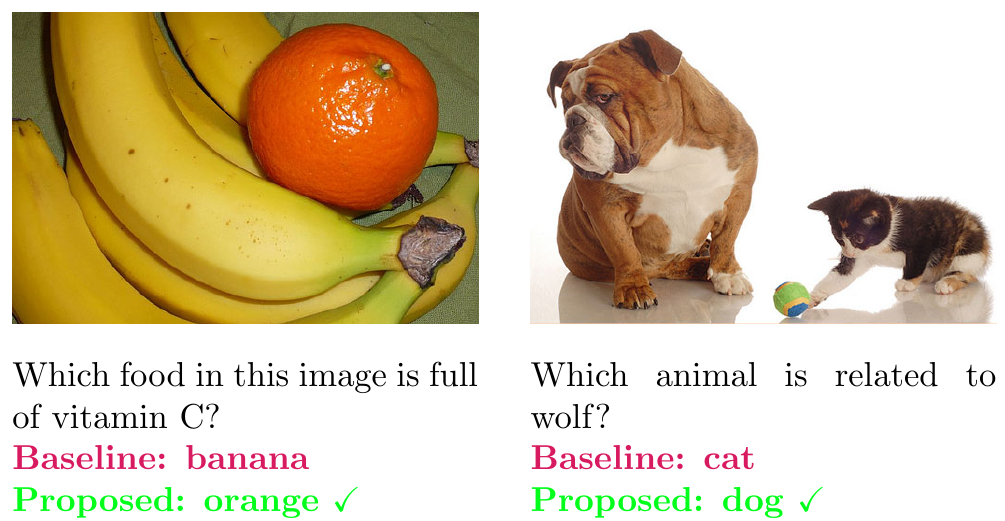}
		\caption{FVQA.\label{fig:fvqa}}
	\end{subfigure}\vspace{4pt}\\
	\begin{subfigure}[b]{0.49\textwidth}
		\centering\includegraphics[width=\textwidth]{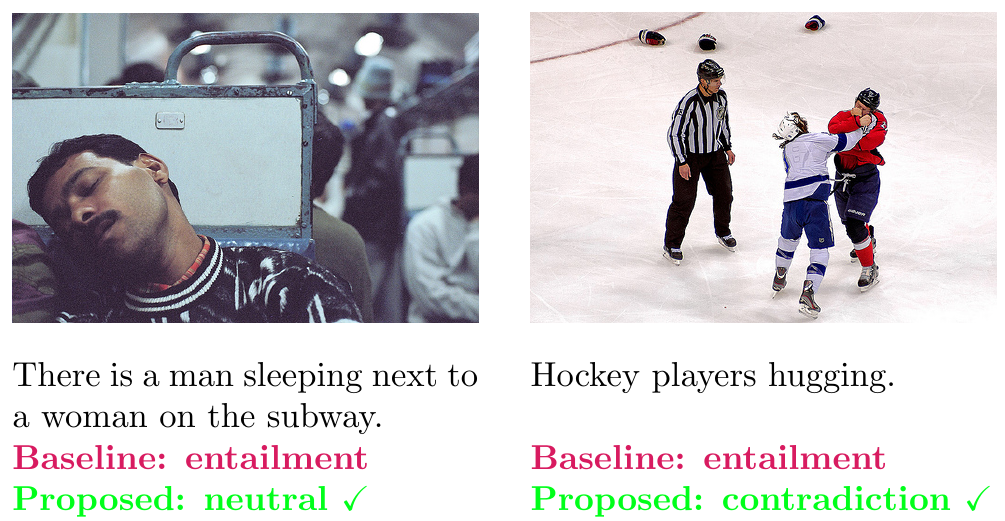}
		\caption{SNLI-VE.\label{fig:snli-ve}}
	\end{subfigure}
	\hfill
	\begin{subfigure}[b]{0.49\textwidth}
		\centering\includegraphics[width=\textwidth]{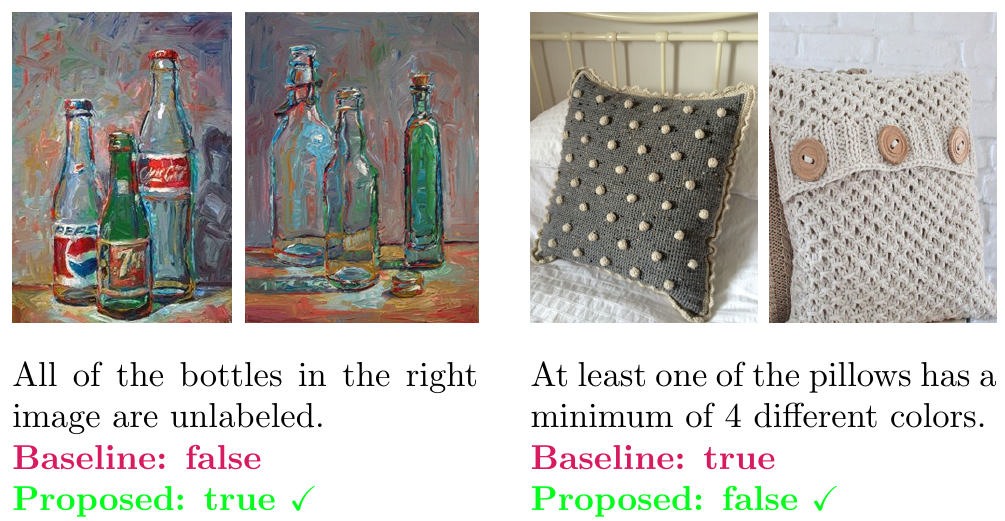}
		\caption{NLVR2.\label{fig:nlvr2}}
	\end{subfigure}
	\vspace{-3pt}
	\caption{Test cases on which our model (Pretraining with VQA v2, GQA w/ ConceptNet PT+FT) produces better predictions than the baseline.
	On OK-VQA and FVQA, the questions require factual knowledge of the type found in ConceptNet.
	On SNLI-VE and NLVR2, the baseline frequently produces common, but incorrect answers.
	See \ap\ref{appendix:results} for additional results including failure cases.}
	\label{fig:examples}
\end{figure*}

\subsection{Results on OK-VQA}
We examine the accuracy on question categories of OK-VQA in \tab\ref{tab:qtypes}.
Each category corresponds to a type of knowledge required (see \ap\ref{appendix:datasets} for a description).
We get \textbf{high gains on categories that correspond to a type of knowledge covered in ConceptNet} (objects properties and features, behavior of people, \etc).
These include OMC (objects, material and clothing), PEL (people and everyday life), BCP (brands, companies and products) and VT (vehicles and transportation).
The only category with a drop in accuracy is ST (science and technology), which is also the smallest (84 questions).
The category with the largest gain is GHLC (geography, history, language, and culture), but it contains only 141 questions, and some are distant from these topics (e.g. \textit{What fruit come from these trees~?}).
These results should not be over interpreted because of the small size of these categories.
Some questions also have imprecise labels. For example, the question \textit{What activity are they doing~?} is labeled with the correct answer \textit{video game}, and our model's answer \textit{play video game} is considered incorrect.

The \textbf{hardest questions for our model} are those referring to exact facts and entities such as place names, famous people, or historical dates. Such precise facts are more difficult to represent and recall than ``soft'' commonsense knowledge. For example, the question \textit{What year was this picture taken~?} requires to recognize a specific event and fetch a precise related fact.
Other difficult questions refer to precise visual cues, while the text of the question is generic, like~\textit{What language is on the sign~?}, \textit{Can you guess the place shown in this picture~?} or \textit{Which season is it~?}. 
Recent analysis showed that V\&L transformers rely primarily on the textual modality~\cite{cao2020behind,singh2020we}.
Additional mechanisms would be needed to allow the recall of facts solely from visual cues.

\subsection{ConceptNet vs Wikidata}
\label{sec:convswiki}
We now examine the suitability of ConceptNet and Wikidata to the four datasets considered.
\textbf{ConceptNet provides larger improvements than Wikidata on every dataset}.
Wikidata shows improvements on knowledge-driven tasks (OK-VQA and FVQA) but fails to improve over the baseline on the visual reasoning ones (SNLI-VE, NLVR2).
Wikidata is almost ten times larger than ConceptNet, but it contains more redundant and noisy information due to its  open-source nature.
ConceptNet, in contrast, is based on a collection of mostly curated sources.
Finally, the vector representations of ConceptNet used were obtained through an advanced and proven procedure that involves ConceptNet as well as other pretrained word representations.
The representations of Wikidata used are a more direct representation of the knowledge graph.

\subsection{Comparison with Existing Methods}

In \tab\ref{tab:comparison} we compare our results with the top entries from the leaderboards of OK-VQA\footnote{\url{https://okvqa.allenai.org}} and NLVR2\footnote{\url{http://lil.nlp.cornell.edu/nlvr/}}.
On OK-VQA, the best accuracy is shown by ConceptBERT model. The other reported results are from the traditional VQA models BAN~\cite{kim2018bilinear} and MUTAN~\cite{ben2017mutan}. The BAN and MUTAN models supplemented with ArticleNet~\cite{marino2019okvqa} obtain each a small improvement (+.44 and +1.43\%). This component retrieves Wikipedia articles from which it extracts an answer for each question. These models perform much worse than the LXMERT baseline, which is trained on multiple datasets.
We include an LXMERT model provided by its authors (LXMERT--Paper) and one retrained with code they provide (LXMERT--GitHub). The latter uses a simplified training strategy, hence a slight discrepancy (\eg 69.50\% on VQA v2 with their model and 68.52\% with the retrained one). Our model brings a clear improvement over LXMERT-github, but it does not surpass LXMERT-paper that we could not reproduce.

On NLVR2, the classical method FiLM~\cite{perez2018film} expectedly performs worse than transformers pretrained on multiple datasets. The LXMERT baseline surpasses VisualBERT~\cite{li2019visualbert}, and our knowledge injection brings a small improvement. The state of the art on NLVR2 is obtained by UNITER~\cite{chen2019uniter} which is pretrained on a much greater amount of captioning data and uses a significantly larger architecture.


\begin{table}[t!]
	\ra{1.15}
	\begin{center}
	\renewcommand\tabcolsep{0pt}
	\footnotesize
	\begin{tabularx}{\linewidth}{Xccc}
		\toprule
		& OK-VQA & ~~~ & NLVR2\\
		\midrule
		MLP & 20.67 && -- \\
		BAN & 25.17 && -- \\
		BAN + ArticleNet & 25.61 && -- \\
		MUTAN & 26.41 && -- \\
		MUTAN + ArticleNet & 27.84 && -- \\
		ConceptBERT & 33.66 && -- \\
		FiLM & -- && 52.1 \\
		VisualBERT \scriptsize{(COCO captions)} & -- && 67.00\\
		UNITER \scriptsize{(COCO, VG, Conceptual Cap., SBU)} & -- && \textbf{79.50}\\
		LXMERT-Paper \scriptsize{(cannot be reproduced, see text)} & \textbf{42.94} && \underline{74.50}\\
		\midrule
		\multicolumn{4}{l}{This paper: ~~\scriptsize{(COCO and VG captions, VQA v2, GQA, VG QA)}}\\
		LXMERT-GitHub & 37.26 && 71.31 \\
		LXMERT-GitHub with ConceptNet & \underline{39.04} && 72.59\\
		\bottomrule
	\end{tabularx}
	\caption{Comparison with existing methods. Datasets used for pretraining are given in parentheses.}
	\label{tab:comparison}
	\end{center}
\end{table}

\subsection{Knowledge Ablation}
The main source of improvements on knowledge-demanding tasks like OK-VQA is the representation of knowledge relevant to test questions.
To illustrate this claim, we create a small knowledge test with OK-VQA to examine how removing certain pieces of knowledge affects model performance.
We select, by keyword search, a small subset of 19 test questions that focus on nutrition, on which our model obtains an accuracy of 91.11\% \textit{vs} 71.11\% for the baseline.
We then identify all entities related to nutrition (\eg \textit{health benefit}, \textit{fiber}, \textit{protein}, \textit{vitamin}, \etc) and remove them from the knowledge base.
After retraining our model with the pruned KB, the performance on nutrition questions drops markedly to 68.89\%. The overall accuracy (on mostly non-nutrition-related questions) is maintained.
This confirms that the withheld knowledge was indeed responsible for the high performance on related questions.
The automated construction of diagnostic tests of this type could allow a quantitative evaluation and would be an interesting direction for future work.


\subsection{Discussion and Limitations}
Our experiments showed that KBs can expand the domain of applicability of V\&L models.
The tested approach proved robust in a range of settings with ConceptNet but the larger Wikidata did not fulfill our expectations.
Our results suggest that the noise and ambiguities in Wikidata prevent realizing its full potential.
Methods for better text-to-knowledge matching such as named entity recognition and homonym disambiguation are promising solutions to investigate.

We also identified the grounding of knowledge with visual evidence to be a limitation currently for certain tasks.
Questions in OK-VQA about specific places or famous people for example require the model to recall specific facts on the basis of precise visual cues.
Although such facts are stored in Wikidata, the tested model did not prove effective at recalling them.
Improvements of the visual grounding could also help with visual reasoning tasks like SNLI-VE where the correct interpretation of a text input is heavily dependent on the visual input.


\section{Conclusions}

In this paper we described a general-purpose technique to inject external information from knowledge bases into multimodal transformers for vision-and-language tasks.
The current prevailing paradigm is to pretrain large models on collections of datasets.
Our experiments demonstrate that some types of commonsense and factual knowledge are not captured within these models. Knowledge bases like ConceptNet and Wikidata can fill in these deficiencies.
We showed clear improvements in performance on a variety of tasks and benchmarks requiring visual and multimodal reasoning, demonstrating the versatility of the procedure.

The value of these results for future research are twofold.
On the one hand, they indicate that the combination of heterogeneous sources of information is a promising way to expand the applicability of current machine learning models.
On the other hand, by improving the availability of supporting knowledge, the approach opens the door to future advances on reasoning procedures that process this information.
Advances on this front would lead to improved capabilities on tasks that require high-level or multi-hop reasoning, for example.


\bibliography{anthology,eacl2021}
\bibliographystyle{acl_natbib}

\clearpage

\appendix

\section*{Reasoning over Vision and Language:\\Exploring the Benefits of Supplemental Knowledge}

\vspace{2pt}
\section*{Appendices}
\vspace{12pt}

\section{Implementation Details}
\label{appendix:implem}

Most transformer-based V\&L models map the input text tokens to word, segment, and position embeddings, that are ultimately combined. Our approach is applied on the word embeddings.

The implementation of our method builds on top of the official implementation of LXMERT\footnote{\url{https://github.com/airsplay/lxmert}} \cite{tan2019lxmert}.
We use values of hyperparameters recommended in the code, including a reduced number of 12 training epochs (compared to 20 mentioned in the paper) and single-stage training strategy.

The loss weight $\lambda$ was selected by cross-validation for maximum accuracy on the OK-VQA validation set.
From the set $\{0.1, 0.5, 1, 10, 100, 200\}$, the optimal values were found to be $0.5$ for Wikidata and $100$ for ConceptNet. The optimal $\lambda$ seems to vary in inverse proportion with the size of the KB (based on Wikipedia, ConceptNet, and three other KBs used in preliminary experiments and not included in the paper).

The fine-tuning experiments used the same batch size of $32$ and learning rate of $0.00005$ as the original LXMERT implementation.
The number of fine-tuning epochs was adjusted according to the dataset size: $25$ for OK-VQA and FVQA, $8$ for NLVR2 and SNLI-VE.

The pretraining experiment used two NVIDIA Tesla M40 GPUs and took about four days per experiment. To pretrain the model on the full five datasets we used four GPUs and each experiment took up to eight days. 
The fine-tuning experiments used a single GPU and took about 1h for FVQA, 3h for OK-VQA, 10h for NLVR2, and 22h for SNLI-VE.

Since the matching of KB entities with the V\&L training data proceeds by exact string matching, we processed the KB entities with the same WordPiece tokenizer~\cite{wu2016google} as the LXMERT model does for the V\&L training data.
To enable fast indexing of the KB, we store the knowledge embeddings $\bV$ as a hash table indexed by textual expressions $\bW$.

\section{Negative Results}

We explored a variety of architectural choices, but some of them did not gain any improvement over the baseline. To measure the distance between learned projections and their embeddings in~\eq\ref*{eq:loss} we tried two other options: smooth L1 loss and cosine distance, but they showed worse results than the used mean squared error. To obtain representations for knowledge-rich expressions we first included additional knowledge embedding layer as a fourth component of textual representations along with word, segment and position embeddings. This technique proved to be inefficient, so eventually we used word embeddings as a target for knowledge alignment.

Further, we tried to leverage information about objects detected in the image. We used predicted labels for each image object as a knowledge-rich expression and added another learning objective to align visual embeddings with corresponding knowledge. With this objective the model converged to a smaller loss during pretraining but the accuracy on all fine-tuning tasks was lower than the baseline results thereby indicating possible overfitting. We think that a better way to exploit visual information would be to use object labels as additional supervision to enhance entity matching.

\section{Linguistic Probing Analysis}
\label{appendix:probe}
We analyzed the representations learned by our model with probing tasks using the SentEval toolkit~\cite{conneau2018senteval}. This complements the evaluation on downstream tasks presented in the main paper.

The probing tasks aim to identify the linguistic information encoded in the learned representations. Following~\cite{cao2020scene}, we tested our model on the following ten linguistic probing tasks~\cite{conneau2018you}:       

\setlist{leftmargin=*,itemsep=2pt}
\begin{itemize}
	\item \textbf{SentLen:} predict the length of sentences;
	\item \textbf{WC:} word content task requires to predict which words appear in the sentence;  
	\item \textbf{TreeDepth:} categorize the sentence according to the depth of its syntax tree; 
	\item \textbf{TopConst:} predict the sequence of top constituents;
	\item \textbf{BShift:} check whether two random adjusted words have been inverted in the sentence;
	\item \textbf{Tense:} predict the tense of the main-clause verb;
	\item \textbf{SubjNum:} predict if the main-clause subject is in singular or plural form;
	\item \textbf{ObjNum:} same as SubjNum but for the main-clause object;
	\item \textbf{SOMO:} semantic odd man count task is to determine if a random noun or verb has been replaced.  
	\item \textbf{CoordInv}: check whether two coordinate clauses have been inverted.
\end{itemize}

These tasks are designed to evaluate the quality of sentence embeddings, but LXMERT model learns separate embeddings for each token. To obtain sentence representations, we thus take outputs of 9 intermediate layers in the language encoder and average each of them across all tokens. Every layer output is used as a separate embedding and we report the best result across all layers (\tab\ref{tab:senteval}). Since the first 9 layers perform attention over textual modality only, no image input is required.

The results show that our model surpasses the baseline in most of the tasks. The highest gain (+9.36\%) is seen in WC category meaning that our model better captures the content of words in a sentence. Importantly, best results for this task are obtained with the first layer outputs. WC accuracy drops for every subsequent layer output til it reach 16.11\% and 18.65\% for the baseline and our model respectively. It implies that information about individual words is well encoded in early layers but gets washed off with further self-attention. Surprisingly, we observe a noticeable gain in accuracy for predicting tense and plurality of words (Tense, SubjNum and ObjNum).
Slight improvements in BShift and CoordInv tasks may be caused by additional constrains imposed on the word order by our model where the order of tokens that constitute a knowledge-rich expression is important. Finally, SentLen and TreeDepth probes highly rely on length of sentences and since both tested models limit textual input to 20 tokens, the results may not be reliable.

\begin{table*}[h!]
	\ra{1.3}
	\begin{center}
	\renewcommand\tabcolsep{3pt}
	\footnotesize
	\begin{tabular}{@{}lcccccccccc@{}}
		 \toprule
		 & SentLen & WC & TreeDepth & TopConst & BShift & Tense & SubjNum & ObjNum & SOMO & CoordInv\\
		 \midrule
		 Baseline & 69.10 (6) & 56.83 (1) & 32.71 (7) & 67.04 (4) & 66.19 (9) & 74.38 (6) & 76.38 (8) & 75.85 (7) & \textbf{51.01 (7)} & 59.52 (6) \\
		 Ours & \textbf{70.36 (5}) & \textbf{66.19 (1)} & \textbf{33.35 (5)} & \textbf{71.49 (6)} & \textbf{68.23 (8)} & \textbf{82.10 (1)} & \textbf{80.26 (7)} & \textbf{79.02 (7)} & 50.70 (5) & \textbf{60.24 (7)} \\
		 \bottomrule
	\end{tabular}
	\caption{Results on linguistic probing tasks. For each, we report the maximum score obtained across layers, with the corresponding layer number in parentheses. Our model (ConceptNet PT) outperforms the baseline (LXMERT) in most probing tasks.}
	\label{tab:senteval}
	\end{center}
\end{table*}

\section{Nearest Neighbors in Embedding Space}
\label{appendix:embspace}

We provide in \tab\ref{tab:neighbors} a random selection of words and their nearest neighbors in the space of learned embeddings. We computed these neighbors using the L2 distance. Using the cosine distance gives qualitatively similar results.

\begin{table*}[h!]
	\ra{1.1}
	\begin{center}
	\renewcommand\tabcolsep{13pt}
	\footnotesize
	\begin{tabular}{@{}lll@{}}
		\toprule
		Word & Nearest neighbors with the baseline & Nearest neighbors with our model\\
		\midrule
		birth & \#\#gm, dat, sensitive, incorporated & births, childbirth, born, 	newborn \\
		boxers & dare, indo, briefs, aiding & boxer, briefs, underwear, panties \\
		creeping & goalscorer, fertility, ineffective, [PAD] & crept, creep, crawling, sneaking \\
		dependent & bacterial, [PAD], \#\#idad, outlaws & depended, depend, depends, dependency \\
		displaced & neptune, roche, peterborough, norway & displacement, refugees, relocated, atletico \\
		down & up, MASK, \#\#combe, \#\#ending & up, on, out, \#\#s \\
		equity & [PAD], implementations, eurasian, newfound & investors, investments, investor, investment \\
		ghosts & [PAD], germain, combustion, \#\#ignment & ghost, ghostly, haunted, phantom \\
		indication & neptune, musee, converting, legion & indications, indicating, signaled, indicative \\
		limb & stump, branch, limbs, thorn & limbs, branch, leg, \#\#wara \\
		policemen & cowboys, firefighters, youths, 37 & policeman, police, cops, constabulary \\
		quebec & sutton, [PAD], monasteries, frederick & montreal, laval, ontario, sudbury \\
		smells & [PAD], aiding, preston, quentin & smelled, smell, odor, scent \\
		successes & [PAD], kilometres, tina, marne & success, achievements, accomplishments, successful \\
		sugar & yeast, powder, memo, coating & chocolate, butter, candy, celaena \\
		taste & feel, smell, fade, become & tastes, flavor, tasted, flavors \\
		unmarried & [PAD], [unused285], [unused685], \#\#dium & divorced, childless, 	widowed, marrying \\
		vintage & retro, antique, victorian, rustic & antique, retro, old, \#\#60 \\
		\bottomrule
	\end{tabular}
	\caption{Selection of words and four nearest neighbors in the space of word embeddings learned by the LXMERT baseline and our model. The embedding space of our model better reflects lexical and semantic similarity.}
	\label{tab:neighbors}
	\end{center}
\end{table*}

\section{Results on NLVR2}
We examine the extended metrics on NLVR2 in \tab\ref{tab:nlvr2-acc}.
The ``consistency'' reflects whether the model answers a given question correctly for all related pairs of images.
Our model shows a \textbf{clear improvement in consistency} over the baseline (34.65 $\rightarrow$ 37.56).
This suggests that the model can better relate a given textual input to different image contexts.
We also report the accuracy on the balanced and unbalanced test sets, designed to evaluate a model's reliance on visual biases.
In the balanced set, every image pair appears twice, one with each label (true/false).
A drop in performance from the standard test set (``All'') to the balanced set (``Bal.'') would indicate that the method exploits biases. Neither the baseline nor any of the described models show an undesirable reliance on image biases.

\begin{table}[b!]
	\ra{1.1}
	\begin{center}
	\renewcommand\tabcolsep{3pt}
	\footnotesize
	\begin{tabularx}{\linewidth}{Xcccc}
		 \toprule
		 & \multicolumn{4}{c}{NLVR2 Test-P}\\
		 \cmidrule{2-5}
		 & All & Consist. & Bal. & Unbal.\\
		 \midrule
		 Baseline (LXMERT) & 71.31 & 34.65 & 70.34 & 72.45 \\
		 \midrule
		 With ConceptNet PT & 71.61 & 34.84 & 71.11 & 72.60 \\
		 With ConceptNet FT & 70.82 & 34.42 & 69.86 & 72.35 \\
		 With ConceptNet PT+FT & \textbf{72.59} & \textbf{37.56} & \textbf{72.12} & \textbf{73.71} \\
		 \bottomrule
	\end{tabularx}
	\vspace{-2pt}
	\caption{Detailed metrics on NLVR2: consistency (\%) and accuracy (\%) on {bal}anced and {unbal}anced subsets. The method brings a clear improvement in consistency.}
	\label{tab:nlvr2-acc}
	\end{center}
\end{table}

\section{Description of Datasets}
\label{appendix:datasets}

\textbf{OK-VQA}~\cite{marino2019okvqa} is an open-ended VQA dataset where all questions require some sort of outside knowledge. It comprises about 14,000 questions about images from the MS COCO~\cite{lin2014microsoft} dataset. All questions are produced by human annotators based either on information found in Wikipedia, or commonsense knowledge and visual evidence from the images. The questions are divided into ten categories (see \tab\ref{tab:qtypes}) according to the type of knowledge needed to answer them. OK-VQA is one of the most diverse VQA datasets currently available that requires general knowledge. We use it accordingly as a primary benchmark in this study.

\textbf{FVQA}~\cite{wang2017fvqa} is a VQA dataset that contains about 5,000 questions that probe for commonsense knowledge. The questions are produced by annotators in procedure that forces the question to involve facts found in a reference KB. Each question in the dataset is therefore associated with one specific ``supporting'' fact that describes a relation between concepts in the question and/or image. We do not use the annotations of these supporting-facts. FVQA provides five different training/test splits. We report results that correspond to the average across the five splits.
The quality of the questions in FVQA is mediocre in comparison to OK-VQA, and it is also much smaller.
We include it in this study because it previously served to evaluate other models designed to use KBs for VQA.

\textbf{SNLI-VE}~\cite{xie2019visual} is a dataset for a visual entailment task. The task is an extension of the classical task of natural language inference (NLI). The visual version involves an image ``premise'' and a text ``hypothesis''. The model must determine whether the hypothesis contradicts the information shown in the image, entails it, or whether there are not enough clues to draw any conclusion. The SNLI-VE dataset contains about 560,000 instances, which were constructed from captions from SNLI~\cite{bowman2015large} paired with images from Flickr30k~\cite{young2014image}.
Despite similarities in the skills required for this task and for VQA, existing VQA models show relatively poor performance on SNLI-VE.
The authors of the dataset attribute it to the need for more fine-grained visual understanding and reasoning, and they suggested the use of external knowledge to improve performance, hence its inclusion in this study.

\textbf{NLVR2}~\cite{suhr2018corpusa} is a dataset that evaluates visual reasoning over pairs of images. Each of the $\sim$107,000 instances in the dataset consists of two images with a statement in natural language.
The model must predict whether the statement accurately describes the pair of images.
The creation of the dataset emphasized the linguistic diversity of the sentences, with the objective for the task to require some compositional reasoning.
We use this task in our study to evaluate the suitability of our model to perform compositional reasoning on the knowledge injected in them from KBs.

\begin{table*}[h!]
	\ra{1.1}
	\begin{center}
	\renewcommand\tabcolsep{13pt}
	\footnotesize
	\begin{tabular}{@{}lll@{}}
		 \toprule
		 Alias & Category & Example\\
		 \midrule
		 VT & Vehicles and Transportation & What is the title of the person driving this vehicle~?\\
		 BCP & Brands, Companies and Products & Name the laptop model shown in this picture~?\\
		 OMC & Objects, Material and Clothing & What sort of room is this woman sitting in~? \\
		 SR & Sports and Recreation & What is this baseball player doing~?\\
		 CF & Cooking and Food & Which of the foods here have the highest saturated fats~?\\
		 GHLC & Geography, History, Language and Culture & What city is this meeting taking place~?\\
		 PEL & People and Everyday Life & What kind of hairstyle does the woman in the black shirt have~?\\
		 PA & Plants and Animals & What sound does this animal make~?\\
		 ST & Science and Technology & Which rodent has a similar name to the technical device seen~?\\
		 WC & Weather and Climate & What kind of clouds are shown~?\\
		 \bottomrule
	\end{tabular}
	\caption{Categories of questions in the OK-VQA dataset corresponding to the type of knowledge required.}
	\label{tab:qtypes}
	\end{center}
\end{table*}

\begin{table*}[h!]
	\ra{1.3}
	\begin{center}
	\renewcommand\tabcolsep{9pt}
	\footnotesize
	\begin{tabular}{@{}lcccc@{}}
		 \toprule
		 & Images & Questions & Answers & Type\\
		 \midrule
		 OK-VQA & 14,031 & 14,055 & 14,456 & knowledge-based VQA\\
		 FVQA & 2,190 & 5,826 & 954 & knowledge-based VQA\\
		 SNLI-VE & 31,783 & 565,286 & 3 & entailment\\
		 NLVR2 & 141,480 & 107,292 & 2 & reasoning\\
		 \bottomrule
	\end{tabular}
	\caption{Summary of datasets used in the experiments.}
	\label{tab:datasets}
	\end{center}
\end{table*}

\clearpage
\section{Additional Qualitative Results}
\label{appendix:results}

\hspace{.5\textwidth}%
\begin{wrapfigure}{l}{\textwidth} 
	\centering
    \begin{subfigure}[t]{1\linewidth}
        \centering\includegraphics[width=\linewidth]{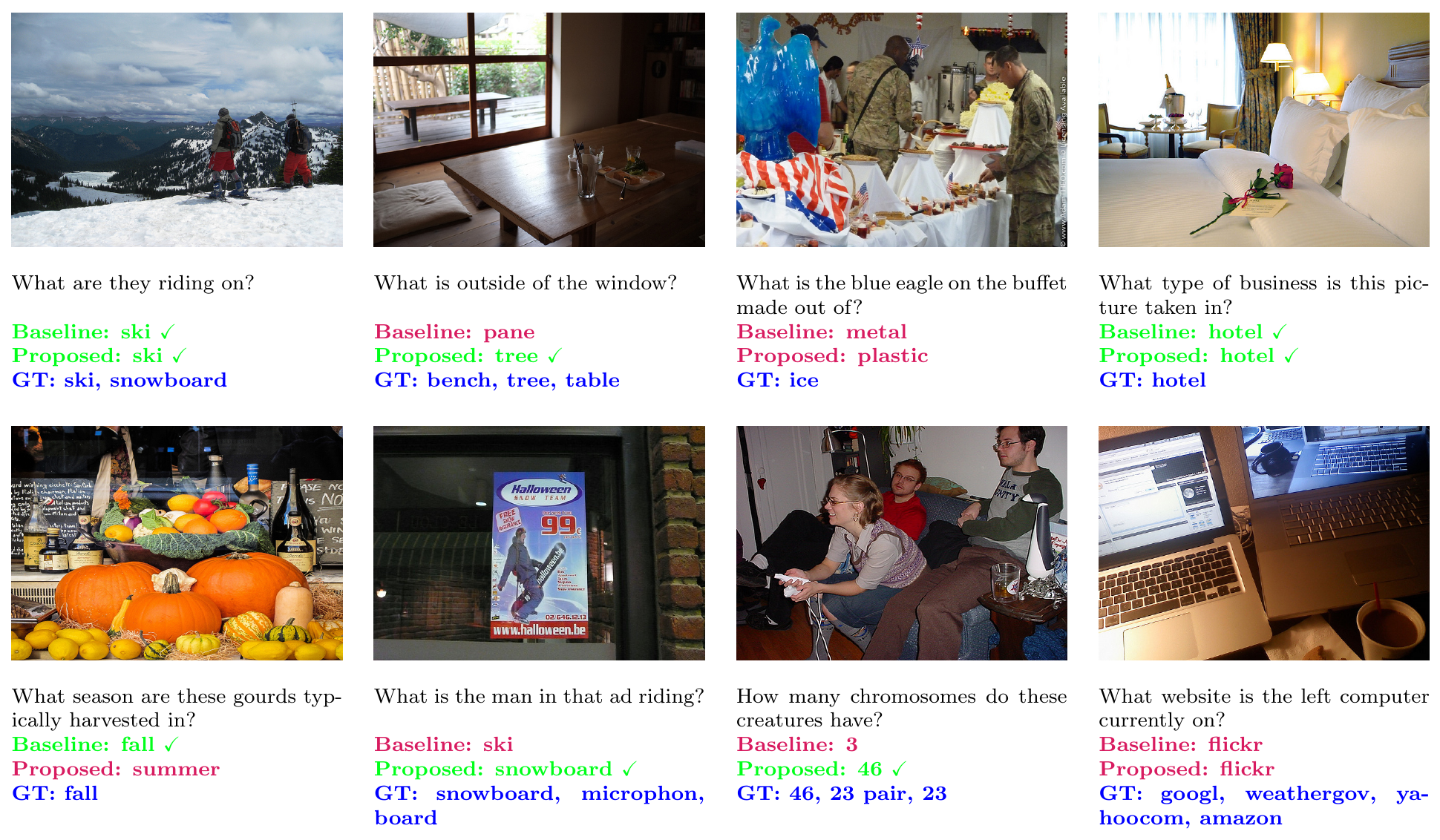}
        \vspace{-26pt}\caption{OK-VQA.}
    \end{subfigure}\vspace{12pt}
    \begin{subfigure}[t]{1\textwidth}
        \centering\includegraphics[width=\linewidth]{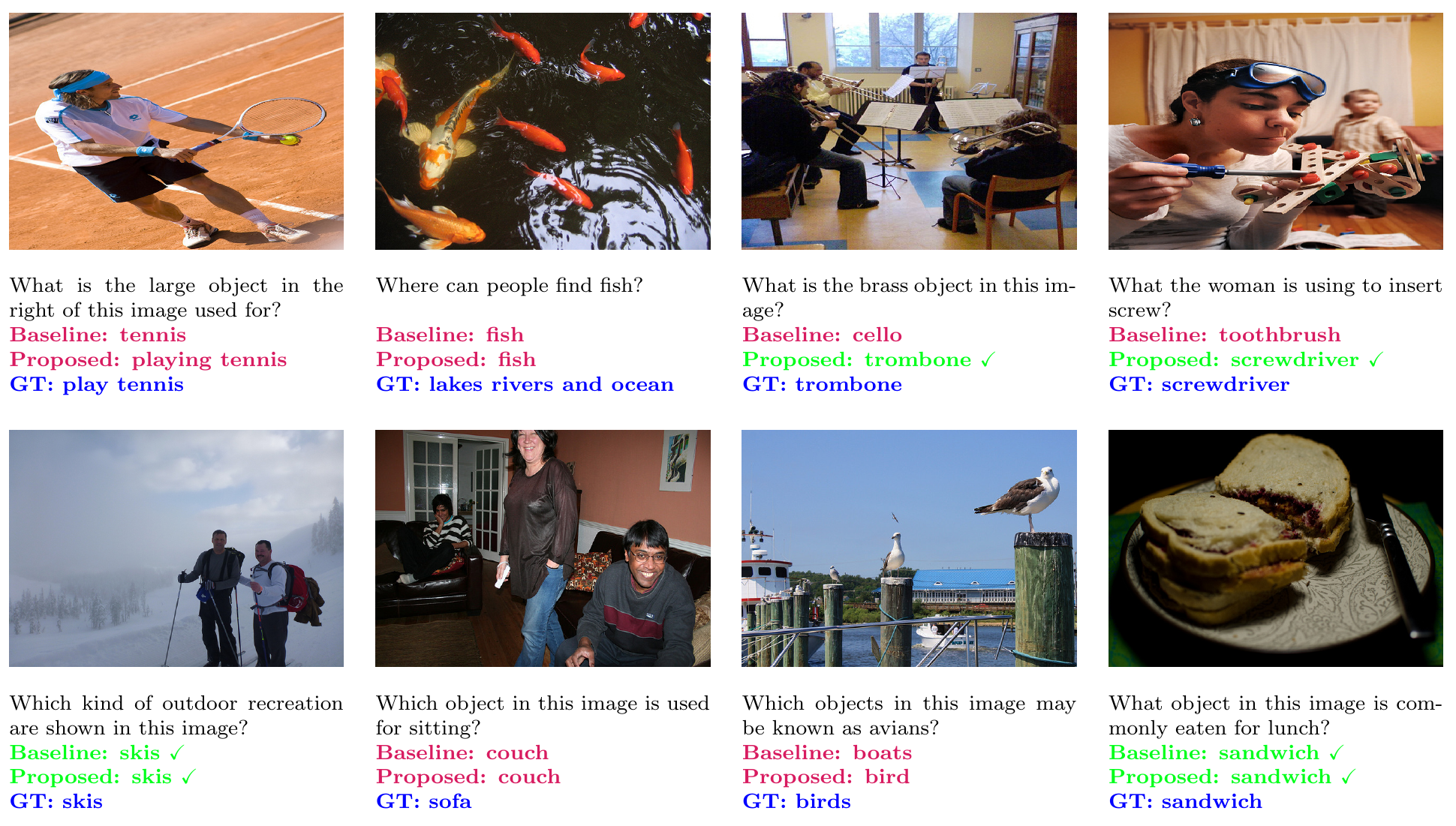}
        \vspace{-18pt}\caption{FVQA.}
    \end{subfigure}\vspace{12pt}
    \caption{Random selection of test instances from OK-VQA and FVQA datasets, with predictions of the baseline (LXMERT) and of our model (Pretraining with VQA v2, GQA w/ ConceptNet PT+FT).}
    \label{fig:more-examples}
\end{wrapfigure}

\begin{figure*}[h!]
	\centering
    \begin{subfigure}[t]{1\textwidth}
        \centering\includegraphics[width=\linewidth]{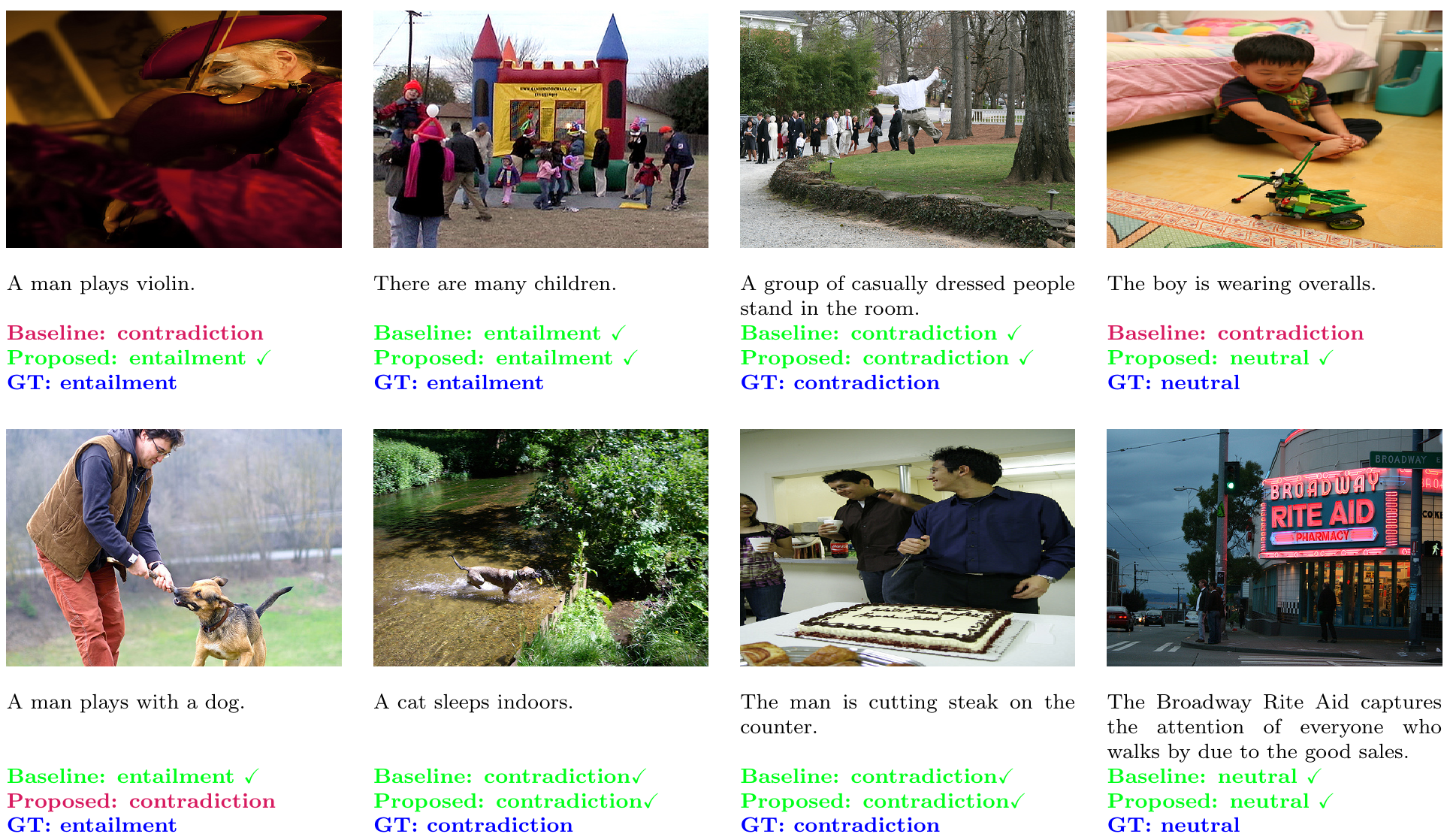}
        \vspace{-18pt}\caption{SNLI-VE.}
    \end{subfigure}\vspace{12pt}
    \begin{subfigure}[t]{1\textwidth}
        \centering\includegraphics[width=\linewidth]{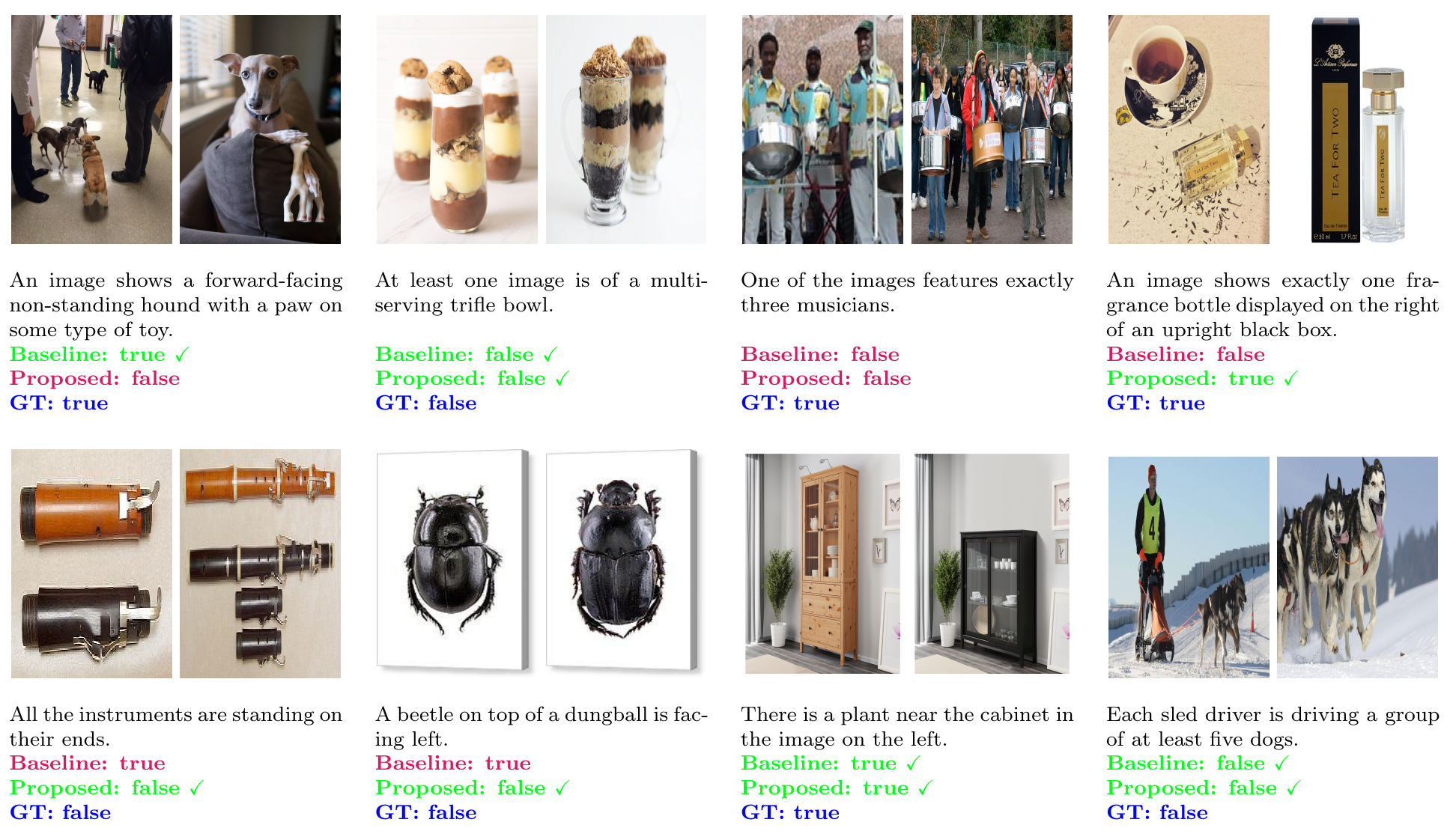}
        \vspace{-18pt}\caption{NLVR2.}
    \end{subfigure}\vspace{12pt}
    \caption{Random selection of test instances from SNLI-VE and NLVR2 datasets, with predictions of the baseline (LXMERT) and of our model (Pretraining with VQA v2, GQA w/ ConceptNet PT+FT).}
    \label{fig:more-examples2}
\end{figure*}

\end{document}